%% file: acl_latex.tex
\documentclass[11pt]{article}

\usepackage[preprint]{acl}

\usepackage{times}
\usepackage{latexsym}
\usepackage[T1]{fontenc}
\usepackage[utf8]{inputenc}
\usepackage{microtype}
\usepackage{inconsolata}
\usepackage{graphicx}
\usepackage{amsmath}
\usepackage{amssymb}
\usepackage{booktabs}
\usepackage{multirow}
\usepackage{subcaption}
\usepackage{xcolor}
\usepackage{xspace}

\input{tables/tab_macros}
\title{WASP: Continual Weight Adaptation for\\ Streaming LLM Personalization}

\author{
  \textbf{Linxi Xie\textsuperscript{1}}\thanks{Work done while interning at Tencent.},
  \textbf{Chonghua Liao\textsuperscript{2,4}},
  \textbf{Mengxuan Wu\textsuperscript{3,4}},
  \textbf{Jiaxin Lin\textsuperscript{4}},
\\
  \textbf{Guandao Yang\textsuperscript{1}},
  \textbf{Zhangyang Wang\textsuperscript{1}},
  \textbf{Kai Wang\textsuperscript{3,4}}
\\
  \textsuperscript{1}University of Texas at Austin,
  \textsuperscript{2}Tsinghua University \\
  \textsuperscript{3}National University of Singapore,
  \textsuperscript{4}Tencent
\\
  \small{\texttt{cissyxie@utexas.edu}}
}

\begin{document}
\maketitle

\begin{abstract}
Personalized large language models (LLMs) should continually adapt to users' evolving preferences. Yet extracting preference changes from noisy interaction histories while supporting inexpensive updates remains challenging. We introduce \textsc{WASP} (\textbf{W}eight \textbf{A}daptation for \textbf{S}treaming \textbf{P}ersonalization), a recurrent framework that captures fine-grained preferences directly from raw user interactions and generates personalized LoRA adapters without expensive per-user optimization. Specifically, \textsc{WASP} achieves this by training a shared hypernetwork that maintains a compact user representation and outputs adapter weights in a single forward pass. As new interactions arrive, \textsc{WASP} updates this representation without reprocessing full interaction history, enabling continual adaptation with per-update costs independent of history length. Experiments on personalization benchmarks and chronological interaction streams show that \textsc{WASP} sustains high-quality personalization over long horizons under fixed compute and storage budgets. Together, these results support efficient streaming personalization and facilitate privacy-preserving on-device deployment. Our project page is available at \url{https://lafoucc.github.io/WASP/}.
\end{abstract}

\input{sec/introduction}

\input{sec/related_work}
\input{sec/preliminary}

\input{sec/method}
\input{sec/experiments}
\input{sec/results}

\input{sec/ablation}
\input{sec/conclusion}

\input{sec/limitations}

\bibliography{custom}

\appendix
\input{sec/appendix}

\end{document}

%% file: tables/tab_macros.tex
\providecommand{\best}[1]{\textbf{#1}}       %
\providecommand{\sbest}[1]{\underline{#1}}   %

%% file: sec/introduction.tex
\section{Introduction}
\label{sec:intro}
\suppressfloats[t]
\begin{figure}[t]
    \centering
    \includegraphics[width=\linewidth]{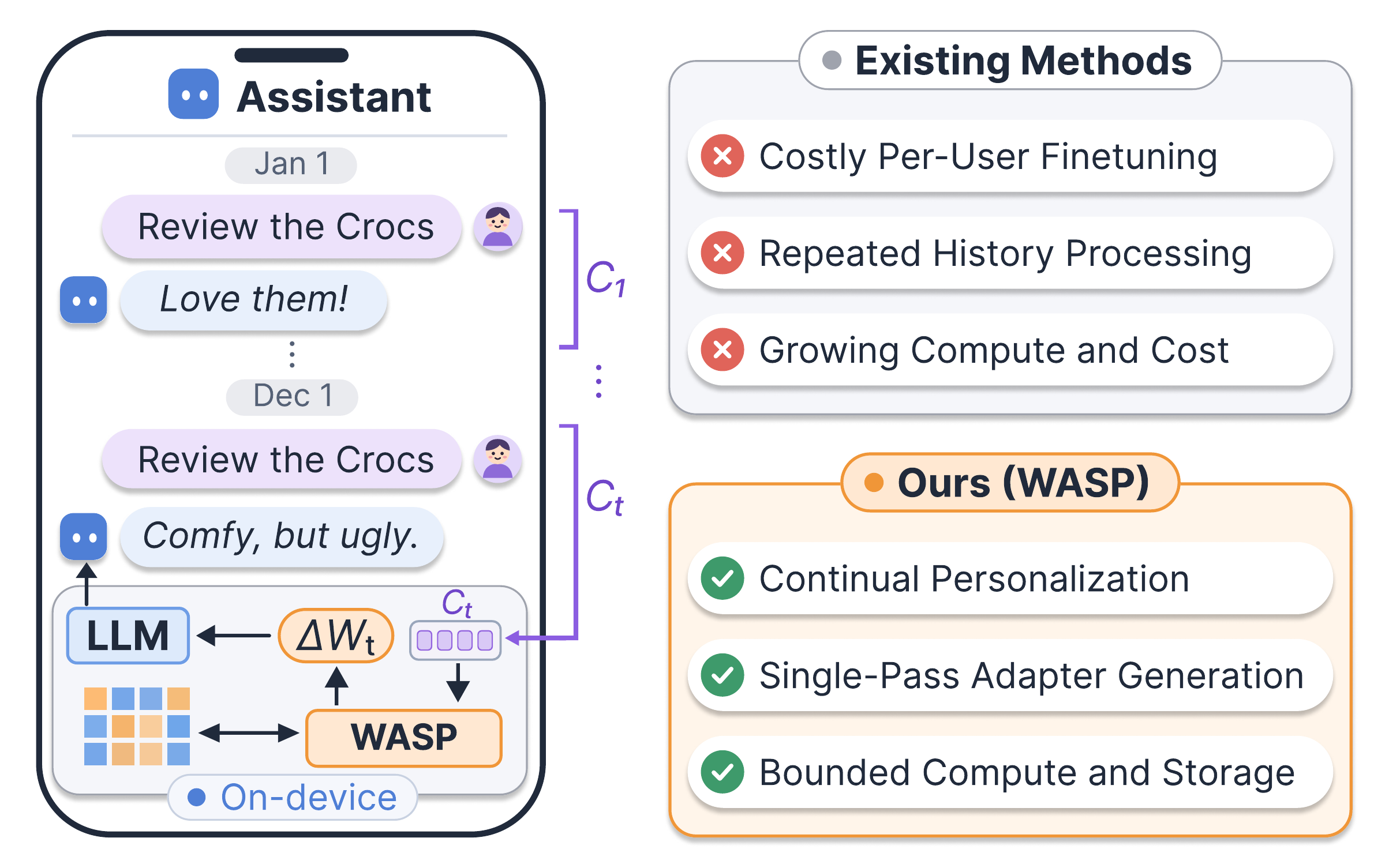}
    \caption{\textsc{WASP} tracks evolving preferences on device without uploading user histories or fine-tuning per user. For each new chunk~$C_t$, \textsc{WASP} updates a compact user state and generates a LoRA adapter~$\Delta W_t$.}
    \label{fig:application}
\end{figure}

Large language models (LLMs) have demonstrated remarkable generative capabilities, yet their one-size-fits-all nature limits their ability to meet the needs and preferences of individual users. LLM personalization addresses this limitation by using a user's interaction history to tailor model responses to their preferences~\cite{salemi2024lamp,personalizationsurvey}. However, tracking evolving preferences through long, noisy interaction histories is challenging because relevant signals are often implicit and scattered across many interactions \cite{personamemv2,springcl}. Moreover, interaction histories may contain sensitive information, making it desirable to keep them on the user’s device. 
Practical personalization must therefore capture fine-grained preferences while supporting inexpensive updates as histories grow. Meeting these requirements would enable efficient streaming personalization and facilitate on-device deployment, allowing sensitive user data to remain local \cite{personalizationsurvey,p2p} (Figure~\ref{fig:application}).

Existing personalization methods often condition generation on full user histories, retrieved interactions \cite{kumar2024longlamp}, or summarized profiles \cite{PAG}. As histories grow, processing this historical context increases inference latency and consumes valuable context window capacity. Moreover, as new interactions arrive, these methods must repeatedly reprocess the accumulated history to incorporate new user evidence. Alternatively, some methods encode user preferences into lightweight, user-specific parameters \cite{tan2024oppu,proper}. While this reduces the need to include histories in the context, conventional per-user fine-tuning requires separate optimization and parameter storage for every user, making it costly to support large numbers of users. Updating these parameters also requires retraining, making continual adaptation costly as preferences evolve \cite{p2p}. Such training can incur substantial latency on resource-constrained devices, while offloading it to the cloud raises concerns about exposing sensitive user data \cite{wang2025ondevice}. Together, these limitations expose a central challenge: how can a personalized model incorporate new user evidence while retaining useful past information without repeatedly reprocessing the full interaction history?

To address this challenge, we propose \textsc{WASP}, a recurrent framework that represents each user with a fixed-size set of parameter-structured latent tokens. These tokens persist across updates and serve as a compact user state that captures the user’s evolving preferences. At each update step, a hypernetwork uses the current user state to attend directly to token-level representations of newly arrived interactions. This allows the framework to extract new evidence about the user and incorporate it into the state. The updated tokens then form the user state for the next step and are decoded into personalized LoRA adapter weights \cite{hu2022lora} in a single forward pass. In this way, the latent tokens retain useful information from past interactions, incorporate new evidence, and generate an updated personalized adapter without reprocessing the entire interaction history. This recurrent design supports effective personalization over a long interaction horizon under fixed compute and storage budgets, facilitating on-device deployment that keeps sensitive user histories local.

We evaluate \textsc{WASP} on LaMP \cite{salemi2024lamp}, LongLaMP \cite{kumar2024longlamp}, and chronological user interaction streams~\cite{amazon2023,tang2008arnetminer}. Across these settings, \textsc{WASP} achieves the strongest average classification performance, remains competitive on long-form generation, and generalizes to unseen users. More importantly, \textsc{WASP} adapts effectively to incoming interactions through recurrent updates and maintains stable performance as histories grow. 
With a fixed-size user state and bounded update costs, \textsc{WASP} supports efficient, continual personalization suitable for resource-constrained on-device deployment.

\begin{figure*}[t]
    \centering
    \includegraphics[width=\textwidth]{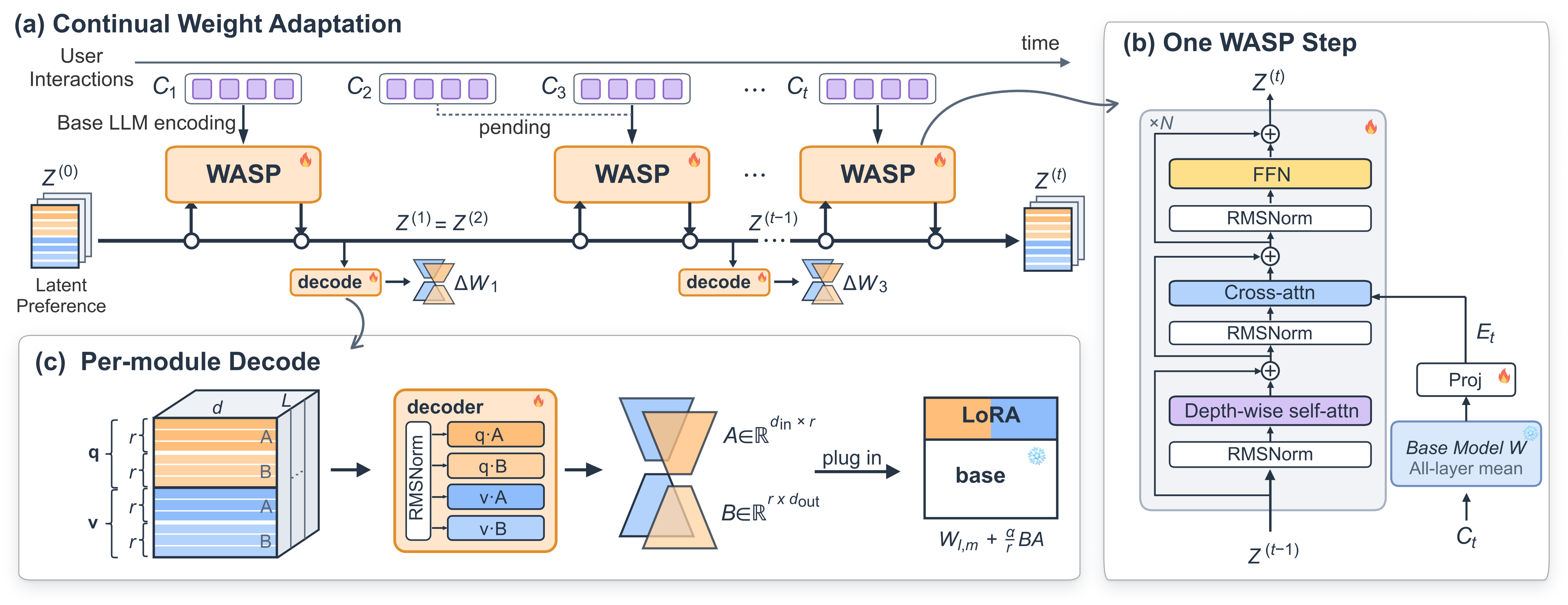}
    \caption{\textsc{WASP} generates updated LoRA weights from a fixed-size user state by processing only the incoming interactions at each step. (a) Parameter-structured LoRA tokens carry the user state across updates. (b) Each \textsc{WASP} step combines inter-layer self-attention with cross-attention to new evidence. (c) Module-specific heads decode the updated tokens into LoRA adapters.}
    \label{fig:pipeline}
\end{figure*}

%% file: sec/related_work.tex
\section{Related Work}
\label{sec:related}

\paragraph{LLM Personalization.}
LLM personalization leverages users' interaction histories to infer their preferences and tailor model responses~\cite{salemi2024lamp,personalizationsurvey}. Context-based methods retrieve relevant past interactions or summarize histories into textual profiles, which are included in the model's input to guide generation~\cite{kumar2024longlamp,PAG,salemi2024optimizing,qiu2025dpl}. These approaches can be further enhanced through RL-based reasoning and user modeling~\cite{wu2025rlpf,amirizaniani2026pr2,zhao2025rlpa,wan2025curiosity,nam2026plus,salemi2025restpg}. Alternatively, parameter-based approaches encode user preferences in user-specific adapters, group-level LoRAs, or representation interventions~\cite{tan2024oppu,tan2024personalizedpieces,proper,liu2026perfit,p2p}. While these methods capture user heterogeneity, efficient adaptation over growing interaction histories remains challenging.

\paragraph{Continual Personalization.}
Continual personalization tracks preference drift while retaining relevant past information~\cite{horizonbench}. Parameter-update methods constrain forgetting through regularization~\cite{kirkpatrick2017ewc,wang2023olora}, discount outdated feedback (NS-DPO), localize LoRA edits (RAIE), or distill follow-up user messages (SDPO)~\cite{nsdpo,zeng2026raie,kleinebuening2026sdpo}. SPRInG combines selective fine-tuning with replay retrieval, while TAP-PER optimizes compact user prefixes~\cite{springcl,beyondretrieval}. These approaches still incur gradient-based adaptation costs. Alternatively, AI Persona revises textual profiles, and PAHF updates preference records from clarification and corrective feedback~\cite{wang2024aipersona,liang2026pahf}. These methods avoid weight updates but still rely on stored textual context to condition generation. These limitations motivate amortized adaptation that encodes evolving preferences directly in model parameters.

\paragraph{Context-to-Parameter Adaptation.}
Hypernetworks generate model parameters from context to amortize adaptation~\cite{ha2017hypernetworks}. P2P maps summarized user profiles to personalized LoRAs~\cite{p2p}, while task-conditioned hypernetworks generate adapters from task embeddings, descriptions, demonstrations, or unlabeled prompts~\cite{mahabadi2021hyperformer,charakorn2025text2lora,phang2023hypertuning,dnd}. Doc-to-LoRA and SHINE distill encoded context into adapters using Perceiver and Transformer hypernetworks, respectively~\cite{doc2lora,shine}, while HY-WU uses structured parameter tokens for conditional weight generation~\cite{hywu}. Most of these methods compile context snapshots without explicitly modeling how the context evolves over time. In continual settings, updating snapshot adapters can require reprocessing accumulated history. Generative Adapter instead accumulates context-derived weight updates in a fixed-size additive state without an explicit forgetting gate~\cite{generativeadapter}, while Doc-to-LoRA and SHINE-R concatenate chunk-level adapters, increasing effective rank and per-user storage~\cite{doc2lora,shine}. These update and storage costs limit their suitability for streaming personalization, motivating a fixed-size recurrent preference state that supports selective revision.

%% file: sec/preliminary.tex
\section{Preliminaries}

\subsection{Problem Formulation}
\label{sec:prelim:prob}

We study LLM personalization in a continual learning setting. For a user $u$, let $\mathcal{H}_{u,t} = ((q_{u,i}, a_{u,i}))_{i=1}^{t-1}$ denote the interaction history preceding step $t$, with $q_{u,i}$ and $a_{u,i}$ representing the query and response
at step $i$. Given a new query $q_{u,t}$, the objective is
to generate a response $a_{u,t}$ aligned with the user's
current preferences, using only evidence available prior
to step $t$.

Following parameter-based personalization methods~\cite{tan2024oppu,p2p}, we encode user preferences in a LoRA adapter $\Delta W_{u,t}$ applied to a shared, frozen backbone $W$, defining the response distribution $p(a_{u,t} \mid q_{u,t}; W \oplus \Delta W_{u,t})$, where $\oplus$ denotes adapter application.
We address two challenges:
(1) mapping raw query--response sequences to effective adapters that capture fine-grained preference signals; and
(2) enabling the adapter to evolve online, integrating new evidence while retaining relevant past information, at a cost independent of history length for a fixed amount of new data.

\subsection{Multi-task Adapted Base Model}
\label{sec:prelim:base}

To separate personalization gains from those due to learning task formats, we first adapt the pretrained backbone $W_0$ to the target tasks, following OPPU~\cite{tan2024oppu}. We train a shared LoRA adapter $G$ on pooled query--response pairs from training users across all target tasks, without conditioning on user histories, by minimizing
$\mathcal{L}(G) = -\mathbb{E}_{(q,a)\sim\mathcal{D}_{\mathrm{train}}}
[\log p(a \mid q; W_0 \oplus G)]$. We merge the learned adapter $G^*$ into $W_0$ to obtain $W$, hereafter the \emph{base model}, which remains frozen throughout personalization training. More details in Appendix \ref{sec:appendix:task-adapted}.

%% file: sec/method.tex
\section{Method}
\label{sec:method}

\textsc{WASP} maps a user's interaction history to a personalized adapter through a fixed-size set of latent tokens. These tokens persist across updates as a compact user state. We introduce the token representation (Section~\ref{sec:method:latent}) and the hypernetwork that generates adapter parameters (Section~\ref{sec:method:arch}), then describe how the user state is updated as new interactions arrive (Section~\ref{sec:method:continual}).

\subsection{Latent Preference Representation}
\label{sec:method:latent}

Inspired by HY-WU~\cite{hywu}, we represent each user with latent \emph{LoRA tokens} $Z \in \mathbb{R}^{L \times 2M \times R \times d}$, where $L$ is the number of backbone layers, $M$ is the number of adapted modules per layer (e.g., query and value projections), $R$ is the LoRA rank, and $d$ is the token dimension. Each module has two sets of $R$ tokens, corresponding to its LoRA factors $A$ and $B$. The token count is independent of history length.

We initialize the tokens as learnable structural embeddings that encode their positions in the adapter (Figure~\ref{fig:pipeline}). Let $S=2MR$ denote the number of token slots within each layer. The initial token at layer $l$ and slot $s$ is:
\begin{equation}
    Z^{(0)}_{l,s}
    = e^{\mathrm{layer}}_l + e^{\mathrm{slot}}_s,
    \text{ for } s=1,\ldots,S.
\end{equation}
The layer embedding identifies the backbone layer, while the slot embedding identifies a token's position within that layer. Both embedding tables are initialized from a standard Gaussian distribution and learned jointly with the generator. 

\subsection{Architecture}
\label{sec:method:arch}
We use a Transformer-based hypernetwork to map user interaction histories to personalized LoRA parameters. Each block comprises inter-layer self-attention, cross-attention for history conditioning, and a feed-forward network. Module-specific decoding heads then map the refined tokens to LoRA parameters (Figure~\ref{fig:pipeline}).

\paragraph{Inter-layer self-attention.}
To enable information flow across backbone layers, we apply bidirectional self-attention along the layer dimension. 
We view the \emph{LoRA tokens} as
$Z\in\mathbb{R}^{L\times S\times d}$, where each slot $s$ identifies the same  adapter components (module, factor, rank) across layers. At block $i$, each slot is updated as:
\begin{equation}
\begin{aligned}
\widetilde Z^{(i)}_{:,s}
&= Z^{(i)}_{:,s}
 + \operatorname{SelfAttn}\!\left(
    \operatorname{Norm}(Z^{(i)}_{:,s})
   \right)
\end{aligned}
\end{equation}
This requires $\mathcal{O}(SL^2)$ pairwise attention interactions, compared with $\mathcal{O}(L^2S^2)$ for full attention over all \emph{LoRA tokens} \cite{hywu}. We also examine other attention groupings (Section ~\ref{sec:ablation:self-attn}).

\paragraph{User history conditioning.}
We encode the user interaction histories at the token level with the task-adapted backbone to retain fine-grained signals of implicit user preferences. We average each token's representations across backbone layers and project them to the hypernetwork dimension, yielding the conditioning sequence $C$. Each block then applies cross-attention followed by a feed-forward network:
\begin{equation}
\begin{aligned}
    \widehat Z^{(i)}
    &= \widetilde Z^{(i)}
      + \operatorname{CrossAttn}\!\left(
          \operatorname{Norm}(\widetilde Z^{(i)}), C
        \right),\\
    Z^{(i+1)}
    &= \widehat Z^{(i)}
      + \operatorname{FFN}\!\left(
          \operatorname{Norm}(\widehat Z^{(i)})
        \right).
\end{aligned}
\end{equation}

\paragraph{Per-module decoding.}
After the final generator block, we map the refined user tokens to adapter parameters using linear decoding heads. Specifically, we define a separate head $\operatorname{Linear}_{m,f}$ for each adapted module type $m$ and LoRA factor $f\in\{A,B\}$:
\begin{equation}
    W_{m,f}
    = \operatorname{Linear}_{m,f}\!\left(
        \operatorname{Norm}(Z_{m,f})
      \right)
\end{equation}
Each head is applied across backbone layers and rank components, with its output width matched to the corresponding backbone dimension. We zero-initialize the $B$ heads so that the generated adapter initially produces no update to the backbone.

\paragraph{End-to-end training.}
We train the hypernetwork end-to-end using the downstream task loss, without requiring pre-optimized user adapters as supervision. Given a history $\mathcal H$, the hypernetwork produces
$\Delta W=\textsc{WASP}_{\phi}(\mathcal H)$. Let $\ell_p$ denote
the response loss under $W\oplus\Delta W$ and $\ell_W$ the loss under
the frozen task-adapted base model $W$. We optimize:
\begin{equation}
    \mathcal L(\phi)
    = \mathbb E\!\left[
        \ell_p+\lambda_{\mathrm{gap}}
        \operatorname{softplus}_{\beta}(\ell_p-\ell_W)
      \right]
\end{equation}
The gap term encourages the generated adapter to improve over the task-adapted base model. Gradients update only the hypernetwork parameters $\phi$.

\subsection{Continual Preference Updates}
\label{sec:method:continual}

Static generation reconstructs an adapter from the available history at
every update. To avoid repeatedly encoding the accumulated history, we
instead carry the latent preference tokens across successive
interactions.

\paragraph{Recurrent generation.}
Let $\mathcal C_t$ denote the interaction chunk received at time step $t$, and let $Z_{t-1}$ be the latent preference state retained from the previous update. We initialize the latent tokens as:
\begin{equation}
    Z_t =
    \begin{cases}
        Z^{(0)}, & t=1,\\
        (1-\Lambda)Z^{(0)}+\Lambda Z_{t-1}, & t>1,
    \end{cases}
\end{equation}
where $\Lambda$ is a learned gate that balances the structural embedding and the previous preference state at each token slot. Starting from this initialization, $Z_t$ attends to the token-level representations of $\mathcal C_t$, allowing the latent state to read and incorporate new preference evidence. The updated state is then decoded into $\Delta W_t$ and carried forward to the next time step.

\paragraph{Surprise-based update.}
Updating after every chunk is unnecessary when the user's preferences remain consistent with the current adapter. For each incoming chunk $\mathcal C_t$, we measure surprise as the difference between its response loss under the current personalized model $W\oplus\Delta W$ and under the task-adapted base model $W$, averaged over the chunk's query--response pairs and normalized to obtain $u_t$. We then accumulate the resulting surprise:
\begin{equation}
    S_t=\max(0,S_{t-1}+u_t),
    \text{ update if } S_t\ge h.
\end{equation}
When an update is triggered, all pending interactions are processed together and $S_t$ is reset. The threshold $h$ controls the trade-off between update frequency and personalization quality.

%% file: sec/experiments.tex
\input{tables/tab_main_random}
\section{Experiments}
\label{sec:exp}

We study \textsc{WASP} in two settings. \emph{Static personalization} tests whether our hypernetwork can effectively map interaction histories to personalized adapters; we therefore evaluate \textsc{WASP}$_{\mathrm{stat}}$ on fixed, non-chronological interaction histories. \emph{Continual personalization} tests whether the recurrent user state can incorporate new evidence while retaining useful past information; we evaluate \textsc{WASP}$_{\mathrm{cont}}$ on chronologically ordered interactions.

\paragraph{Datasets.}
For static personalization, following P2P~\cite{p2p}, we deploy LaMP~\cite{salemi2024lamp} and LongLaMP~\cite{kumar2024longlamp}. LaMP comprises three classification tasks, one rating-prediction task, and three text-generation tasks, while LongLaMP comprises three long-form generation tasks. Each instance contains a user interaction history, a query, and a ground-truth response.

For continual personalization, we study two long-form text-generation tasks: review generation on Amazon Reviews 2023 (Amazon23)~\cite{amazon2023} and temporal abstract generation from LongLaMP (AbsGen)~\cite{kumar2024longlamp}. We order each user's interactions chronologically and partition them into successive chunks. We also evaluate explicit preference following on a semi-synthetic MovieLens dataset with controlled changes in users' genre preferences
(Appendix~\ref{sec:appendix:dataset:movielens}).

\paragraph{Baselines.}
We compare methods that condition on the full user history (\textsc{Full-History}), retrieved interactions (\textsc{RAG}), or summarized profiles (\textsc{PAG}). We also compare parameter-based methods: \textsc{MT-LoRA} fine-tunes one shared adapter on interactions pooled across training users, whereas \textsc{OPPU} fine-tunes a separate adapter on each test user's history \cite{tan2024oppu}. P2P~\cite{p2p} and SHINE~\cite{shine} use hypernetworks to generate fixed adapters from compressed representations of user history. In the efficiency study~\ref{sec:results:efficiency}, we additionally consider their continual variants, \textsc{P2P}$_{\mathrm{reappend}}$ and SHINE-R. All methods use the same task-adapted Qwen2.5-7B-Instruct backbone. Baseline details are provided in Appendix~\ref{sec:appendix:baselines}.

\paragraph{Evaluation Settings.}
For static personalization, we follow P2P~\cite{p2p} and evaluate on both \emph{random} and \emph{out-of-distribution} (\textsc{OOD}) user splits. The random split samples test users to reflect the overall user distribution, while the OOD split selects users whose profiles differ most from those of the training users. The latter provides a more challenging test of generalization to unseen users.

For continual personalization, we partition each user's interactions into chronologically ordered chunks $\mathcal C_{u,1},\ldots,\mathcal C_{u,T}$, each containing a sequence of query--response pairs. When evaluating on $\mathcal C_{u,t+1}$, methods may access only preceding chunks $\mathcal C_{u,1:t}$. We average performance first across chunks for each user, then across users within each task, preventing users with longer histories from disproportionately affecting the results.

\paragraph{Evaluation metrics.}
We report accuracy and F1 for classification, MAE and RMSE for rating prediction, and ROUGE-1 and ROUGE-L for text generation. In the continual setting, we additionally report BLEU and METEOR for text generation. To assess efficiency and scalability, we measure update and generation FLOPs, per-user storage, and peak memory. Full implementation and hyperparameter details are provided in Appendix~\ref{sec:appendix:implementation}.

%% file: tables/tab_main_random.tex
\begin{table*}[t]
\centering
\small
\setlength{\tabcolsep}{4pt}
\renewcommand{\arraystretch}{1.05}
\resizebox{\textwidth}{!}{%
\begin{tabular}{ll cccc cc ccc}
\toprule
\multirow{2}{*}{\textbf{Task}} & \multirow{2}{*}{\textbf{Metric}}
  & \multicolumn{4}{c}{\textbf{Context-based}}
  & \multicolumn{2}{c}{\textbf{PEFT-based}}
  & \multicolumn{3}{c}{\textbf{Hypernetwork-based}} \\
\cmidrule(lr){3-6}\cmidrule(lr){7-8}\cmidrule(lr){9-11}
 & & Base Model & Full hist. & RAG & PAG & MT-LoRA & OPPU & SHINE & P2P & Ours \\
\midrule
\multirow{2}{*}{LaMP-1 Citation ID}        & Acc                & .524 & .622 & .516 & .543 & .594 & .579 & .610 & \sbest{.689} & \best{.728} \\
                                           & F1                 & .507 & .617 & .495 & .530 & .591 & .566 & .606 & \sbest{.689} & \best{.728} \\
\multirow{2}{*}{LaMP-2N News Cat.}         & Acc                & .688 & .796 & .616 & .752 & .712 & \best{.860} & \sbest{.842} & .706 & .829 \\
                                           & F1                 & .477 & .627 & .480 & .581 & .527 & \best{.676} & .661 & .455 & \sbest{.672} \\
\multirow{2}{*}{LaMP-2M Movie Tagging}     & Acc                & .459 & .456 & .446 & .425 & .519 & .506 & \sbest{.537} & .485 & \best{.668} \\
                                           & F1                 & .416 & .406 & .399 & .394 & .486 & .471 & \sbest{.504} & .429 & \best{.638} \\
\multirow{2}{*}{LaMP-3 Product Rating}     & MAE $\downarrow$   & .294 & .258 & .385 & .303 & .285 & .244 & \sbest{.235} & .258 & \best{.217} \\
                                           & RMSE $\downarrow$  & .613 & .583 & .810 & .696 & .620 & \sbest{.563} & \best{.538} & .590 & \best{.538} \\
\multirow{2}{*}{LaMP-4 Headline Gen.}      & R-1                & .174 & .182 & .169 & .164 & \sbest{.183} & .179 & \best{.187} & .182 & .180 \\
                                           & R-L                & .155 & \sbest{.163} & .153 & .148 & .162 & .162 & \best{.167} & \sbest{.163} & .162 \\
\multirow{2}{*}{LaMP-5 Title Gen.}         & R-1                & .491 & \sbest{.503} & \best{.517} & .499 & .490 & .486 & .493 & .492 & .497 \\
                                           & R-L                & .432 & \sbest{.443} & \best{.457} & .440 & .433 & .432 & .433 & .436 & .438 \\
\multirow{2}{*}{LaMP-7 Tweet Paraphrase}   & R-1                & .512 & .521 & .504 & .505 & .521 & .522 & \sbest{.523} & \best{.528} & .518 \\
                                           & R-L                & .450 & .459 & .446 & .449 & .456 & .458 & \sbest{.461} & \best{.465} & .456 \\
\multirow{2}{*}{LongLaMP-1 Abstract}       & R-1                & .384 & .344 & .370 & .370 & \sbest{.387} & .355 & .384 & .326 & \best{.390} \\
                                           & R-L                & .228 & .199 & .219 & .218 & .229 & .218 & \sbest{.230} & .191 & \best{.231} \\
\multirow{2}{*}{LongLaMP-2 Topic}          & R-1                & .290 & .280 & .275 & .280 & .291 & .288 & \sbest{.297} & .286 & \best{.298} \\
                                           & R-L                & .141 & .139 & .138 & .142 & .143 & .143 & .143 & \sbest{.144} & \best{.147} \\
\multirow{2}{*}{LongLaMP-3 Review}         & R-1                & .310 & .317 & .292 & .305 & .305 & .328 & \best{.337} & .313 & \sbest{.335} \\
                                           & R-L                & .156 & .155 & .151 & .155 & .154 & .161 & \best{.166} & .157 & \sbest{.164} \\
\midrule
\multirow{2}{*}{\textbf{Avg.\ classification}} & Acc                & .557 & .625 & .526 & .573 & .609 & .648 & \sbest{.663} & .626 & \best{.742} \\
                                           & F1                 & .467 & .550 & .458 & .502 & .535 & .571 & \sbest{.590} & .525 & \best{.679} \\
\multirow{2}{*}{\textbf{Avg.\ generation}} & R-1                & .360 & .358 & .354 & .354 & \sbest{.363} & .360 & \best{.370} & .354 & \best{.370} \\
                                           & R-L                & .260 & .260 & .261 & .259 & .263 & .262 & \best{.267} & .259 & \sbest{.266} \\
\bottomrule
\end{tabular}%
}
\caption{On the random split of LaMP and LongLaMP, \textsc{WASP}$_{\mathrm{stat}}$ achieves the highest average classification accuracy and F1 and ties for the highest average ROUGE-1. Methods are grouped into context-based, PEFT-based, and hypernetwork-based personalization.}
\label{tab:main-random}
\end{table*}

%% file: sec/results.tex
\input{tables/tab_main_ood}
\section{Results}
\label{sec:results}

\subsection{Static Personalization}
\label{sec:results:static}

\paragraph{Effective personalization across users.}
Tables~\ref{tab:main-random} and~\ref{tab:main-ood} show that \textsc{WASP} achieves the strongest average classification performance across both user splits while remaining competitive on long-form generation. The gains over the task-adapted base model show that the generated adapters capture user-specific information beyond shared task knowledge. Compared with \textsc{Full-History}, \textsc{WASP} improves average classification accuracy/F1 by 11.7/12.9 points on the random split and 8.1/10.4 points on the OOD split, suggesting that more history is not always beneficial. On generation, \textsc{WASP} ranks first or second on every LongLaMP metric across both splits and ties for the highest average ROUGE-1 on the random split.

\paragraph{\textsc{WASP} generalizes across history lengths.}
\input{figures/fig_cold_start}
Within the static setting, we independently generate adapters for 200 users from non-chronological histories of varying lengths (Figure~\ref{fig:cold_start}). \textsc{WASP}$_{\mathrm{stat}}$ remains strongest across all context lengths, demonstrating its ability to map variable-length histories to effective LoRA adapters. Its advantage is clearest beyond $10^3$ tokens, where the baselines plateau or decline. More details are in Appendix~\ref{sec:appendix:cold-start}.

\subsection{Continual Personalization}
\label{sec:results:continual}
\input{tables/tab_continual_main}

\paragraph{Effective continual personalization.}
Table~\ref{tab:continual-main} shows that recurrent updates improve personalization over long-horizon interaction streams.
Compared with methods that repeatedly reprocess the accumulated history—either by appending the full history to the model input (\textsc{Full-History}) or by fine-tuning a new LoRA on it at each update (\textsc{OPPU})—\textsc{WASP}$_{\mathrm{cont}}$ achieves higher ROUGE scores on both tasks while retaining only a fixed-size latent state. 
It also outperforms methods with a static user representation on seven of eight metrics, demonstrating the importance of updating the preference state as user preferences evolve.

\paragraph{Performance remains stable over long horizons.}
\input{figures/fig_continual_quality}
Figure~\ref{fig:continual-quality} further examines performance as the accumulated history grows. \textsc{WASP}$_{\mathrm{cont}}$ remains stable across successive updates, while its advantage over the static variant generally widens over time. This trend indicates that the recurrent state continues to incorporate new evidence without progressively forgetting previously observed preferences. On a semi-synthetic MovieLens benchmark, $\mathrm{WASP}_{\mathrm{cont}}$ achieves higher accuracy than an RPMem-style update on both revised preferences (1.000 vs.~.329) and retained preferences for genres omitted from the update (.997 vs.~.898) (Appendix~\ref{sec:appendix:preference-revision}).

\subsection{Efficiency and Scalability}
\label{sec:results:efficiency}
\input{figures/fig_efficiency}

\paragraph{Efficient updates and compact storage.}
The update computation and per-user storage costs of \textsc{WASP}$_{\mathrm{cont}}$ remain independent of accumulated history length (Figure~\ref{fig:efficiency}a--b). By processing only the incoming chunk and maintaining a fixed-size latent state, our method bounds its update cost by the chunk size and keeps storage fixed at 8.7MB per user. In contrast, methods that reprocess accumulated history become increasingly expensive: at 20K tokens, P2P's profile regeneration requires approximately $3.5\times$ as many update FLOPs as ours, while OPPU requires over $100\times$ as many. Storage also grows for \textsc{Full-History}, which retains the history's KV cache, and for SHINE-R, which stores a separate LoRA for each chunk, each exceeding 1\,GB per user at 20K tokens. These bounded resource requirements make \textsc{WASP}$_{\mathrm{cont}}$ well suited to
on-device streaming personalization.

\paragraph{Bounded serving cost.}
\textsc{WASP}$_{\mathrm{cont}}$ keeps both generation cost and peak memory independent of accumulated history length (Figure~\ref{fig:efficiency}c--d). Its fixed-rank adapter matches \textsc{Naive}'s generation cost while requiring only 2.2\,GB of additional peak memory. In contrast, \textsc{Full-History}'s generation cost grows with its cached context, while SHINE-R's accumulated per-chunk LoRAs increase its effective rank and peak memory. P2P incurs the largest memory footprint because its 4B profile encoder must remain resident during adapter regeneration.

%% file: tables/tab_main_ood.tex
\begin{table*}[t]
\centering
\small
\setlength{\tabcolsep}{4pt}
\renewcommand{\arraystretch}{1.05}
\resizebox{\textwidth}{!}{%
\begin{tabular}{ll cccc cc ccc}
\toprule
\multirow{2}{*}{\textbf{Task}} & \multirow{2}{*}{\textbf{Metric}}
  & \multicolumn{4}{c}{\textbf{Context-based}}
  & \multicolumn{2}{c}{\textbf{PEFT-based}}
  & \multicolumn{3}{c}{\textbf{Hypernetwork-based}} \\
\cmidrule(lr){3-6}\cmidrule(lr){7-8}\cmidrule(lr){9-11}
 & & Base Model & Full hist. & RAG & PAG & MT-LoRA & OPPU & SHINE & P2P & Ours \\
\midrule
\multirow{2}{*}{LaMP-1 Citation ID}        & Acc                & .533 & \sbest{.624} & .498 & .600 & .573 & .537 & .514 & .600 & \best{.733} \\
                                           & F1                 & .493 & \sbest{.610} & .495 & .600 & .570 & .493 & .475 & .599 & \best{.731} \\
\multirow{2}{*}{LaMP-2N News Cat.}         & Acc                & .637 & \sbest{.671} & .589 & .578 & .656 & .602 & \best{.696} & .589 & .662 \\
                                           & F1                 & .507 & .541 & .491 & .495 & .537 & .522 & \best{.552} & .414 & \sbest{.542} \\
\multirow{2}{*}{LaMP-2M Movie Tagging}     & Acc                & .513 & .567 & .543 & .533 & .587 & \sbest{.648} & .614 & .516 & \best{.712} \\
                                           & F1                 & .359 & .374 & .353 & .328 & \sbest{.446} & .396 & .404 & .375 & \best{.564} \\
\multirow{2}{*}{LaMP-3 Product Rating}     & MAE $\downarrow$   & \sbest{.217} & .244 & .272 & .240 & .221 & .221 & \best{.194} & .249 & \sbest{.217} \\
                                           & RMSE $\downarrow$  & .611 & .640 & .696 & .630 & .615 & .615 & \best{.517} & .615 & \sbest{.547} \\
\multirow{2}{*}{LaMP-4 Headline Gen.}      & R-1                & .220 & \best{.241} & .221 & .220 & .215 & \sbest{.233} & \best{.241} & .225 & .221 \\
                                           & R-L                & .199 & \best{.221} & .205 & .201 & .196 & .209 & \sbest{.217} & .204 & .200 \\
\multirow{2}{*}{LaMP-5 Title Gen.}         & R-1                & .482 & \sbest{.499} & .490 & \best{.501} & .474 & .484 & .482 & .483 & .465 \\
                                           & R-L                & .432 & \best{.452} & .440 & \sbest{.447} & .424 & .432 & .425 & .434 & .412 \\
\multirow{2}{*}{LaMP-7 Tweet Paraphrase}   & R-1                & .511 & \best{.608} & .563 & \sbest{.570} & .518 & .511 & .555 & .541 & .528 \\
                                           & R-L                & .438 & \best{.550} & .501 & \sbest{.507} & .449 & .441 & .485 & .469 & .456 \\
\multirow{2}{*}{LongLaMP-1 Abstract}       & R-1                & .367 & .343 & .354 & .363 & \sbest{.371} & .351 & .367 & .309 & \best{.383} \\
                                           & R-L                & .218 & .199 & .209 & .214 & \sbest{.220} & .211 & .219 & .181 & \best{.229} \\
\multirow{2}{*}{LongLaMP-2 Topic}          & R-1                & .291 & .278 & .276 & .268 & \sbest{.292} & .289 & \best{.299} & .282 & \sbest{.292} \\
                                           & R-L                & .143 & .141 & .142 & .137 & .143 & \sbest{.144} & \best{.145} & .142 & \sbest{.144} \\
\multirow{2}{*}{LongLaMP-3 Review}         & R-1                & .293 & .298 & .270 & .275 & .291 & .306 & \best{.330} & .288 & \sbest{.326} \\
                                           & R-L                & .144 & .148 & .137 & .139 & .144 & .149 & \best{.159} & .145 & \sbest{.157} \\
\midrule
\multirow{2}{*}{\textbf{Avg.\ classification}} & Acc                & .561 & \sbest{.621} & .543 & .570 & .605 & .596 & .608 & .568 & \best{.702} \\
                                           & F1                 & .453 & .508 & .446 & .474 & \sbest{.518} & .470 & .477 & .463 & \best{.612} \\
\multirow{2}{*}{\textbf{Avg.\ generation}} & R-1                & .361 & \sbest{.378} & .362 & .366 & .360 & .362 & \best{.379} & .355 & .369 \\
                                           & R-L                & .262 & \best{.285} & .272 & .274 & .263 & .264 & \sbest{.275} & .262 & .266 \\
\bottomrule
\end{tabular}%
}
\caption{On the OOD split of LaMP and LongLaMP, \textsc{WASP}$_{\mathrm{stat}}$ leads in average classification accuracy and F1 and ranks in the top two on every LongLaMP metric. Column groups follow Table~\ref{tab:main-random}.}
\label{tab:main-ood}
\end{table*}

%% file: figures/fig_cold_start.tex
\begin{figure}[t]
\centering
\includegraphics[width=\linewidth]{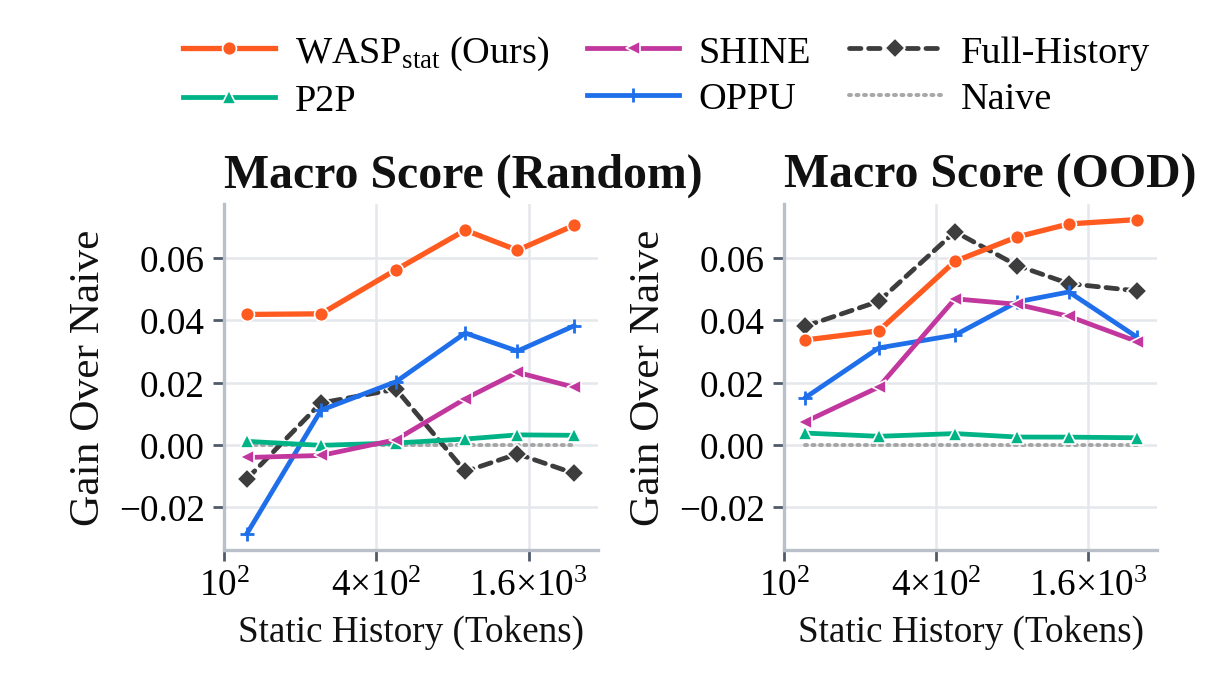}
\caption{\textsc{WASP}$_{\mathrm{stat}}$ achieves the highest macro score across the tested history lengths, with its clearest lead beyond $10^3$ tokens. Scores are averaged across ten tasks for 200 users. Appendix~\ref{sec:appendix:cold-start} defines the macro score.}
\label{fig:cold_start}
\end{figure}

%% file: tables/tab_continual_main.tex
\begin{table*}[t]
\centering
\small
\setlength{\tabcolsep}{2.5pt}
\renewcommand{\arraystretch}{1.05}
\resizebox{\textwidth}{!}{%
\begin{tabular}{llccccccc}
\toprule
\textbf{Dataset} & \textbf{Metric} & \textbf{Base Model} & \textbf{Full hist.}
  & \textbf{OPPU} & \textbf{SHINE} & \textbf{P2P}
  & \textbf{\textsc{WASP}$_{\mathrm{stat}}$} & \textbf{\textsc{WASP}$_{\mathrm{cont}}$ (Ours)} \\
\midrule
\multirow{4}{*}{Amazon23}
  & R-1 $\uparrow$     & .2789$_{\pm.0059}^{\dagger}$ & .2866$_{\pm.0054}^{\dagger}$ & .3263$_{\pm.0060}^{\dagger}$ & \sbest{.3351}$_{\pm.0053}$ & .3099$_{\pm.0061}^{\dagger}$ & .3292$_{\pm.0051}^{\dagger}$ & \best{.3357}$_{\pm.0053}$ \\
  & R-L $\uparrow$     & .1437$_{\pm.0030}^{\dagger}$ & .1494$_{\pm.0027}^{\dagger}$ & \sbest{.1617}$_{\pm.0033}^{\dagger}$ & .1576$_{\pm.0027}^{\dagger}$ & .1542$_{\pm.0028}^{\dagger}$ & \sbest{.1617}$_{\pm.0029}^{\dagger}$ & \best{.1659}$_{\pm.0030}$ \\
  & METEOR $\uparrow$  & .1449$_{\pm.0045}^{\dagger}$ & .1511$_{\pm.0043}^{\dagger}$ & .1798$_{\pm.0051}^{\dagger}$ & \best{.1880}$_{\pm.0045}$ & .1673$_{\pm.0047}^{\dagger}$ & .1822$_{\pm.0044}^{\dagger}$ & \sbest{.1873}$_{\pm.0045}$ \\
  & BLEU $\uparrow$    & .0097$_{\pm.0008}^{\dagger}$ & .0119$_{\pm.0011}^{\dagger}$ & \sbest{.0180}$_{\pm.0019}^{\dagger}$ & .0172$_{\pm.0016}^{\dagger}$ & .0141$_{\pm.0012}^{\dagger}$ & \sbest{.0180}$_{\pm.0014}^{\dagger}$ & \best{.0195}$_{\pm.0017}$ \\
\midrule
\multirow{4}{*}{AbsGen}
  & R-1 $\uparrow$     & .3985$_{\pm.0030}^{\dagger}$ & \sbest{.4029}$_{\pm.0038}$ & .3924$_{\pm.0028}^{\dagger}$ & .3940$_{\pm.0031}^{\dagger}$ & .3981$_{\pm.0032}^{\dagger}$ & .4013$_{\pm.0034}^{\dagger}$ & \best{.4052}$_{\pm.0032}$ \\
  & R-L $\uparrow$     & .2344$_{\pm.0029}^{\dagger}$ & \sbest{.2421}$_{\pm.0037}^{\dagger}$ & .2246$_{\pm.0027}^{\dagger}$ & .2341$_{\pm.0032}^{\dagger}$ & .2418$_{\pm.0032}^{\dagger}$ & .2419$_{\pm.0032}^{\dagger}$ & \best{.2451}$_{\pm.0033}$ \\
  & METEOR $\uparrow$  & .2633$_{\pm.0033}^{\ddagger}$ & \best{.2688}$_{\pm.0041}^{\ddagger}$ & \sbest{.2639}$_{\pm.0029}^{\ddagger}$ & .2546$_{\pm.0037}^{\dagger}$ & .2531$_{\pm.0036}^{\dagger}$ & .2564$_{\pm.0036}^{\dagger}$ & .2599$_{\pm.0036}$ \\
  & BLEU $\uparrow$    & .0595$_{\pm.0021}^{\dagger}$ & \best{.0714}$_{\pm.0029}^{\ddagger}$ & .0553$_{\pm.0018}^{\dagger}$ & .0557$_{\pm.0021}^{\dagger}$ & .0574$_{\pm.0021}^{\dagger}$ & .0580$_{\pm.0021}^{\dagger}$ & \sbest{.0609}$_{\pm.0023}$ \\
\bottomrule
\end{tabular}}
\caption{Recurrent updates significantly improve \textsc{WASP} over its static variant on all four metrics across both datasets. Scores are averaged over chunks per user, then across users, with 95\% user-bootstrap confidence intervals. $\dagger$ and $\ddagger$ mark baseline scores significantly below and above \textsc{WASP}$_{\mathrm{cont}}$, respectively ($p<0.05$).}
\label{tab:continual-main}
\end{table*}

%% file: figures/fig_continual_quality.tex
\begin{figure}[t]
\centering
\includegraphics[width=\linewidth]{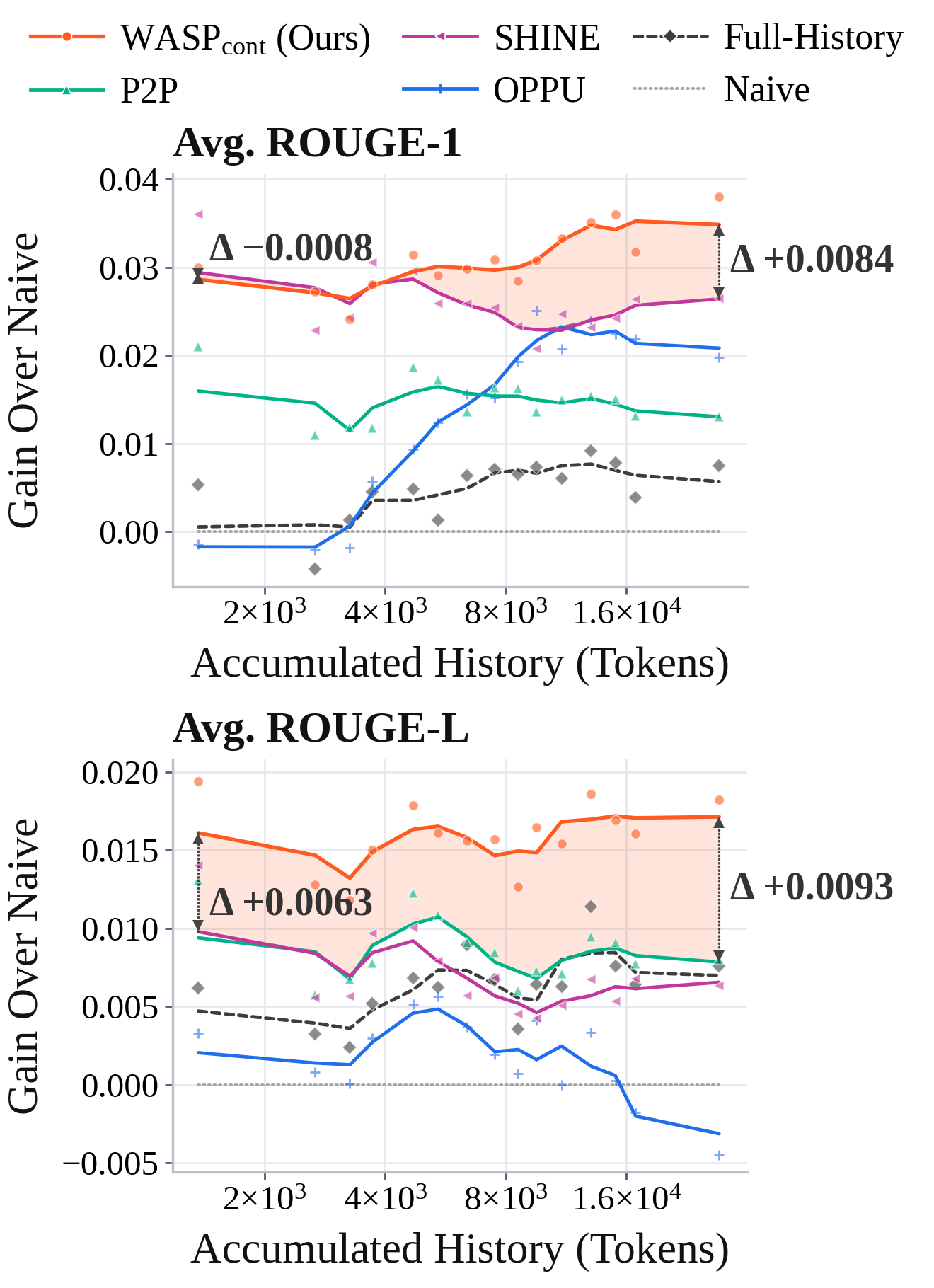}
\caption{\textsc{WASP}$_{\mathrm{cont}}$ maintains stable personalization quality as history grows, and its lead over the static variant generally widens. The curves show average ROUGE-1 and ROUGE-L gains over the unpersonalized base model.}
\label{fig:continual-quality}
\end{figure}

%% file: figures/fig_efficiency.tex
\begin{figure}[t]
\centering
\includegraphics[width=\linewidth]{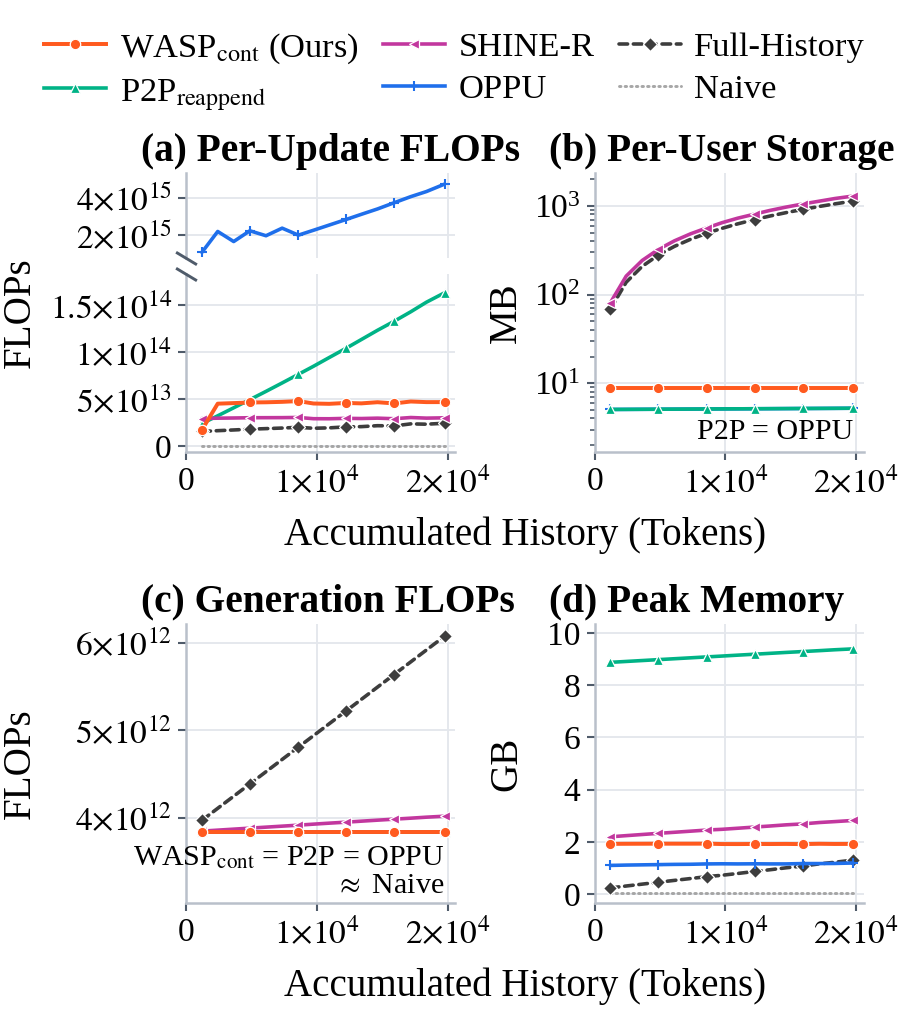}
\caption{\textsc{WASP}$_{\mathrm{cont}}$ keeps computation, storage, and peak memory bounded as user histories grow. Full-History reuses a KV cache of prior interactions.}
\label{fig:efficiency}
\end{figure}

%% file: sec/ablation.tex
\section{Ablation}
\label{sec:ablation}

\begin{table}[t]
    \centering
    \small
    \setlength{\tabcolsep}{4.5pt}
    \begin{tabular}{lcccc}
        \toprule
        \textit{Intra-layer attn.} & Acc. & F1 & R-1 & R-L \\
        \midrule
        Base Model      & .5590 & .4600 & .3605 & .2610 \\
        Intra-full      & .7090 & .6140 & .3590 & .2650 \\
        Intra-rank      & .6905 & .5985 & .3655 & .2660 \\
        No intra (ours) & \textbf{.7220} & \textbf{.6455} & \textbf{.3695} & \textbf{.2660} \\
        \midrule
        \textit{Surprise thr.} & R-1 & R-L & METEOR & BLEU \\
        \midrule
        Base Model & .3387 & .1891 & .2041 & .0346 \\
        $h=0$ & \textbf{.3705} & \textbf{.2055} & \textbf{.2236} & .0402 \\
        $h=0.25$   & .3701 & \textbf{.2055} & .2229 & \textbf{.0405} \\
        $h=1$      & .3687 & .2050 & .2218 & .0398 \\
        \bottomrule
    \end{tabular}
    \caption{Top: Inter-layer attention alone matches or outperforms both intra-layer variants, averaged over the random and OOD splits. Bottom: Updating more often improves personalization; $h=0$ (update every chunk) and $h=0.25$ both outperform $h=1$, averaged over Amazon23 and AbsGen.}
    \label{tab:ablation}\label{tab:intra-attention}\label{tab:surprise-threshold}
\end{table}

\paragraph{Intra-layer attention is unnecessary.}
\label{sec:ablation:self-attn}
Table~\ref{tab:intra-attention} (top) compares our inter-layer self-attention design (\emph{No intra}) with two variants that add intra-layer attention: \emph{Intra-rank} mixes the $R$ rank components while \emph{Intra-full} mixes all $S=2MR$ LoRA-token slots within a layer. Averaged across the random and OOD splits, \emph{No intra} performs best, suggesting that within-layer mixing may blur component specialization. We therefore adopt the simpler and more efficient inter-layer-only design.

\paragraph{Surprise threshold.}
Table~\ref{tab:surprise-threshold} (bottom) compares update thresholds $h$. Lower thresholds trigger more frequent updates and outperform $h=1$ on all four metrics, demonstrating a trade-off between update cost and personalization quality.

%% file: sec/conclusion.tex
\section{Conclusion}
\label{sec:conclusion}

We present \textsc{WASP}, a recurrent framework for streaming LLM personalization. It carries a fixed-size latent user state across updates to incorporate new user evidence and generate personalized LoRA weights without reprocessing accumulated user histories. Trained end-to-end, \textsc{WASP} delivers effective personalization across users and over long interaction horizons while maintaining personalization quality under fixed compute and storage budgets. These results highlight the potential of recurrent weight generation for resource-constrained on-device personalization.

%% file: sec/limitations.tex
\section*{Limitations}
\label{sec:limitations}
We identify two limitations of WASP. First, our evaluation focuses on implicit preferences and includes a preliminary experiment on explicit preference revision (Appendix~\ref{sec:appendix:preference-revision}). Preference following when relevant cues are sparse amid unrelated dialogue is a direction for future study~\cite{personamemv2}. Second, we focus on fixed-rank LoRA adapters in this work. Although the framework could be extended to other parameter-efficient methods, such as adapter modules and prefix tuning, their effectiveness and efficiency in recurrent personalization remain to be evaluated.

%% file: sec/appendix.tex
\section{Additional Results}\label{sec:appendix:results}

\subsection{Generalization Across History Lengths}
\label{sec:appendix:cold-start}
\begin{figure*}[t]
    \centering
    \includegraphics[width=\textwidth]{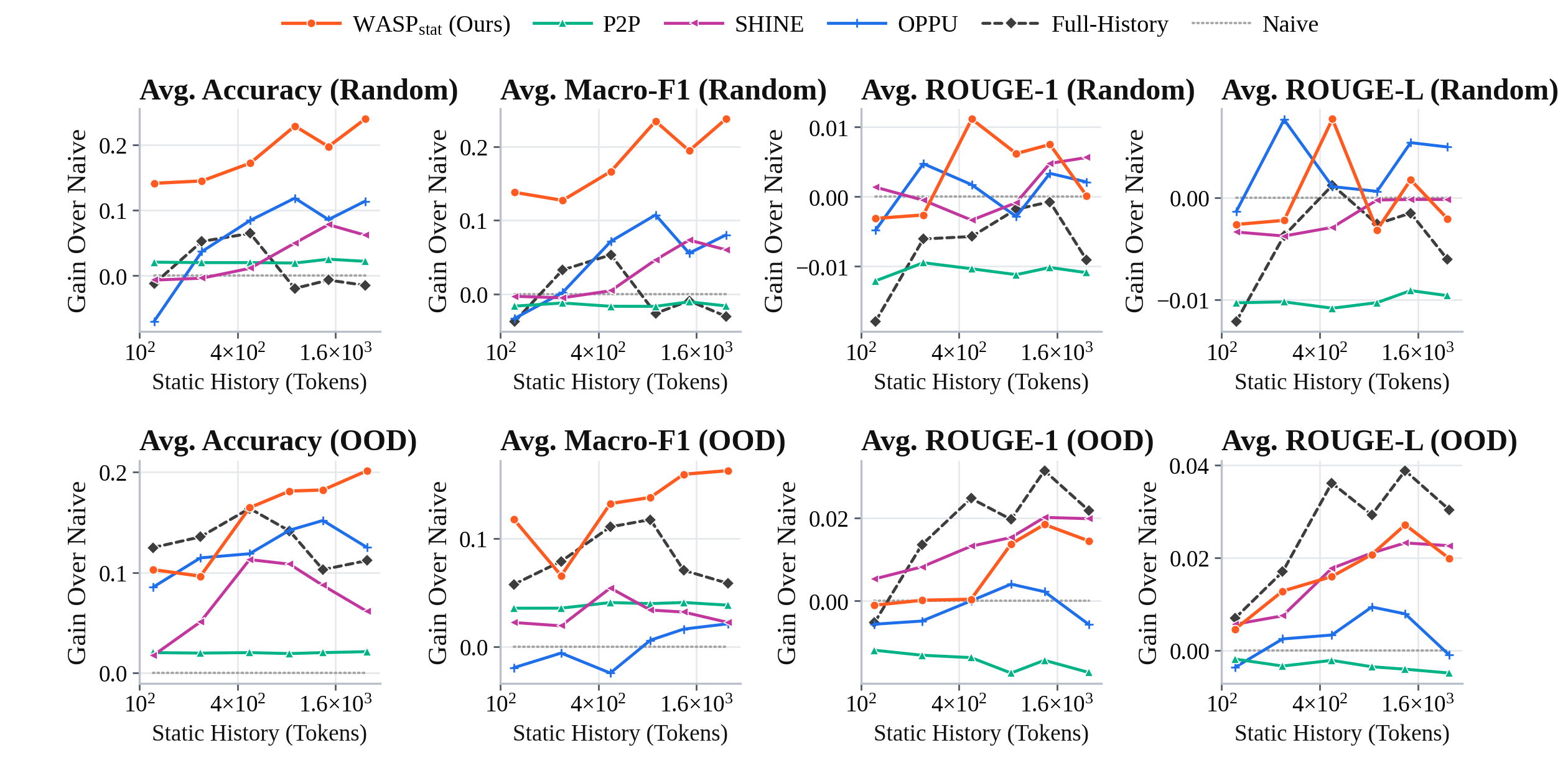}
    \caption{\textsc{WASP}$_{\mathrm{stat}}$ leads on classification and remains competitive on generation across history lengths on both the random and OOD splits. The top row shows the random split, and the bottom row shows the OOD split. The panels report gains in accuracy, macro-F1, ROUGE-1, and ROUGE-L over the task-adapted base model.}
    \label{fig:cold-start-full-metrics}
\end{figure*}

The macro score in Figure~\ref{fig:cold_start} uses one primary metric per task: accuracy for classification, $1-\mathrm{MAE}/4$ for rating prediction, and ROUGE-L for generation, averaged equally across all ten tasks. Figure~\ref{fig:cold-start-full-metrics} reports the corresponding accuracy, macro-F1, ROUGE-1, and ROUGE-L gains over the task-adapted base model for the random and OOD splits. Consistent with Section~\ref{sec:results:static}, these detailed results show that \textsc{WASP}$_{\mathrm{stat}}$ has a clear advantage on classification while remaining competitive on generation across both splits; its overall lead is most pronounced beyond $10^3$ tokens, where it continues to benefit from longer histories as several baselines plateau or decline.

\subsection{User-Specific Personalization}
Strong task performance alone does not necessarily imply effective personalization: a method may learn a population-level adapter without capturing meaningful differences among users~\citep{qiu2025dpl}. We therefore compare performance when conditioning on the target user's history (\emph{self}) against both a shuffled history from another user (\emph{shuffle}) and the task-adapted base model. We define the personalization rate as $\text{win\_both}=\Pr(\text{self}\succeq\text{shuffle}\wedge\text{self}\succeq\text{base})$, which measures how often the correct user history improves over both controls.

\input{tables/tab_personalization}

Table~\ref{tab:winboth} reports the user-specific personalization results on the random split. \textsc{WASP} achieves the highest aggregate personalization rate across all ten tasks, with the strongest F1, ROUGE, and MAE results. To further assess inter-user variation, we measure the cosine similarity between adapters generated for different users. \textsc{WASP} yields the lowest inter-user cosine similarity (0.572), while P2P's near-unity similarity (0.998) indicates that its generated adapters are almost identical across users. Together, these results suggest that \textsc{WASP} transforms fine-grained history signals into distinct and useful user adaptations rather than a largely shared adapter encoding population-level preferences.

\subsection{Targeted Preference Revision}
\label{sec:appendix:preference-revision}

Our continual personalization experiments show that $\mathrm{WASP}_{\mathrm{cont}}$ maintains stable performance over long streams of real-world interactions. Because these interactions reveal preferences only indirectly, it is difficult to determine which preferences the model revises after an update. We therefore use a semi-synthetic MovieLens benchmark with explicit genre preferences and controlled changes between two chunks (Appendix~\ref{sec:appendix:dataset:movielens}).

We compare $\mathrm{WASP}_{\mathrm{cont}}$ with the base model, Full-History, OPPU, and $\mathrm{WASP}_{\mathrm{stat}}$. We also evaluate an RPMem-style update~\cite{rpmem}, which encodes each new chunk independently and merges it with the previous state through a learned gate. After the second chunk, we measure accuracy on held-out movies according to whether the update confirms the user's preference for that genre (\emph{restate}), reverses it (\emph{revise}), or leaves it unmentioned (\emph{retention}).

In $\mathrm{WASP}_{\mathrm{cont}}$, the previous user state attends directly to the new chunk, allowing the network to decide which preferences to revise and which to retain. As Table~\ref{tab:movielens-preference-revision} shows, $\mathrm{WASP}_{\mathrm{cont}}$ follows explicit reversals (1.000 revision accuracy) while retaining preferences for genres absent from the update (0.997 retention accuracy).

\begin{table*}[t]
    \centering
    \small
    \setlength{\tabcolsep}{4.5pt}
    \begin{tabular}{lcccccc}
        \toprule
        Target & Base Model & Full hist. & OPPU & $\mathrm{WASP}_{\mathrm{stat}}$ & RPMem & $\mathrm{WASP}_{\mathrm{cont}}$ \\
        \midrule
        Restate   & .510 & .599 & .783 & \sbest{.993} & .990 & \best{1.000} \\
        Revise    & .513 & .503 & \sbest{.514} & .010 & .329 & \best{1.000} \\
        Retention & .510 & .552 & .647 & \sbest{.993} & .898 & \best{.997} \\
        \midrule
        Overall   & .511 & .551 & .648 & .668 & \sbest{.739} & \best{.999} \\
        \bottomrule
    \end{tabular}
    \caption{On MovieLens, recurrent updates let WASP follow revised preferences and retain those for genres absent from the update.}
    \label{tab:movielens-preference-revision}
\end{table*}

\section{Additional Ablations}

\subsection{Multi-Task Adaptation Phase}
\label{sec:appendix:task-adapted}

We ablate the task-adaptation stage by training \textsc{WASP} directly on the pretrained backbone $W_0$, requiring the hypernetwork to learn task formats and personalization jointly. Table~\ref{tab:task-adapted-phase} compares this variant with representative baselines on the random and OOD splits.

\begin{table*}[t]
    \centering
    \small
    \setlength{\tabcolsep}{4.5pt}
    \begin{tabular}{lllccccc}
        \toprule
        Split & Task & Metric & Base Model & Full hist. & OPPU & P2P & \shortstack{\textsc{WASP}\\w/o TA} \\
        \midrule
        \multirow{4}{*}{Random}
        & \multirow{2}{*}{Avg. classification} & Acc. $\uparrow$ & .505 & \sbest{.615} & .568 & .580 & \best{.712} \\
        & & F1 $\uparrow$ & .496 & .528 & .557 & \sbest{.566} & \best{.645} \\
        & \multirow{2}{*}{Avg. generation} & R-1 $\uparrow$ & .287 & .330 & .301 & \sbest{.332} & \best{.362} \\
        & & R-L $\uparrow$ & .207 & .220 & .221 & \sbest{.244} & \best{.264} \\
        \midrule
        \multirow{4}{*}{OOD}
        & \multirow{2}{*}{Avg. classification} & Acc. $\uparrow$ & .532 & \sbest{.607} & .528 & .581 & \best{.667} \\
        & & F1 $\uparrow$ & .525 & .480 & .507 & \best{.563} & \sbest{.558} \\
        & \multirow{2}{*}{Avg. generation} & R-1 $\uparrow$ & .294 & \sbest{.343} & .305 & .326 & \best{.371} \\
        & & R-L $\uparrow$ & .211 & .232 & .226 & \sbest{.243} & \best{.271} \\
        \bottomrule
    \end{tabular}
    \caption{Without task adaptation, \textsc{WASP} achieves the best result on seven of eight aggregate metrics across the random and OOD splits. \textsc{WASP} w/o TA is trained directly on the pretrained backbone $W_0$.}
    \label{tab:task-adapted-phase}
\end{table*}

Without task adaptation, \textsc{WASP} achieves the best result on seven of eight aggregate metrics, showing that our hypernetwork remains effective without task-adapted initialization. We nevertheless use the task-adapted base model $W$ in the main experiments to disentangle gains from personalization and those from learning task formats.

\subsection{User Representation}
We represent each user history using its layer-averaged hidden states from the task-adapted backbone (Section~\ref{sec:method:arch}). Because this requires passing the raw history through the backbone, we evaluate a cheaper variant that instead encodes the history using Qwen3-Embedding-0.6B~\cite{qwen3}, a separate text encoder. We conduct this comparison on Amazon23~\cite{amazon2023}.

\begin{table}[t]
    \centering
    \small
    \setlength{\tabcolsep}{6pt}
    \begin{tabular}{lccc}
        \toprule
        Metric & Naive & Text enc. & Backbone \\
        \midrule
        R-1 $\uparrow$    & .3012 & \sbest{.3307} & \best{.3342} \\
        R-L $\uparrow$    & .1511 & \sbest{.1636} & \best{.1653} \\
        METEOR $\uparrow$ & .1641 & \sbest{.1860} & \best{.1888} \\
        BLEU $\uparrow$   & .0135 & \sbest{.0192} & \best{.0196} \\
        \bottomrule
    \end{tabular}
    \caption{Encoding user histories with the task-adapted backbone yields the best results on all four Amazon23 metrics, with Qwen3-Embedding-0.6B close behind. Naive is the unpersonalized task-adapted model.}
    \label{tab:user-representation}
\end{table}

As shown in Table~\ref{tab:user-representation}, the backbone representations remain best, suggesting that their richer token-level signals are useful for personalization. Nevertheless, the small gap demonstrates that \textsc{WASP} remains effective with a cheaper, coarser user representation.

\section{Implementation Details}
\label{sec:appendix:implementation}

We use Qwen2.5-7B-Instruct as the backbone. Unless noted otherwise, all LoRA adapters use $r{=}8$, $\alpha{=}8$, and target \texttt{q\_proj} and \texttt{v\_proj}. The \textsc{WASP} hypernetwork has 836M parameters.

\subsection{Static Personalization}
We train a shared task-adaptation LoRA on the training splits of all ten tasks for 4{,}000 steps (batch size 64; learning rate $1\times10^{-4}$) and merge it into the backbone. We then train \textsc{WASP}$_{\mathrm{stat}}$ for 20{,}000 steps (batch size 64; learning rate $1\times10^{-5}$), truncating user histories to 4{,}096 tokens. For classification, we weight each example by $n^{-1/2}$, where $n$ is its label frequency. We set $\lambda_{\mathrm{gap}}{=}0.3$ and $\beta{=}5$ in the gap term $\lambda_{\mathrm{gap}}\operatorname{softplus}_{\beta}(\ell_p-\ell_W)$, where $\ell_p$ and $\ell_W$ denote the response losses of the personalized and frozen task-adapted models, respectively.

\subsection{Continual Personalization}
We train the shared task-adaptation LoRA on the training splits of Amazon23 and AbsGen for 1{,}200 steps (batch size 64; learning rate $1\times10^{-4}$) and merge it into the backbone. Continual \textsc{WASP}$_{\mathrm{cont}}$ is initialized from the static model. We train for up to 4{,}000 steps using 32 episodes per batch, a learning rate of $1\times10^{-5}$, and the same gap loss. All main continual results (Table~\ref{tab:continual-main}) update user state after every chunk ($h=0$). Table~\ref{tab:surprise-threshold} evaluates the surprise trigger (Section~\ref{sec:method:continual}) at $h=0$, $h=0.25$ and $h=1$. Table~\ref{tab:update-counts} reports the number of updates each threshold triggers.

\begin{table}[h]
\centering
\small
\begin{tabular}{lccc}
\toprule
Dataset & $h=0$ & $h=0.25$ & $h=1$ \\
\midrule
Amazon23 & 5.83 (100\%)  & 2.87 (49\%) & 2.12 (36\%) \\
AbsGen   & 11.19 (100\%) & 5.19 (46\%) & 2.89 (26\%) \\
\bottomrule
\end{tabular}
\caption{Mean number of state updates per test user for each surprise threshold.}
\label{tab:update-counts}
\end{table}

\subsection{Computation Resources}
We train on 32 NVIDIA H20 GPUs and run inference on 8 NVIDIA RTX PRO 6000 Blackwell GPUs (96\,GB). We additionally measure the wall-clock latency of one \textsc{WASP}$_{\mathrm{cont}}$ update (batch size 1) on a single NVIDIA GH200 and a single RTX PRO 6000, averaged over each user's updates and then across users (Table~\ref{tab:update-latency}). The backbone and hypernetwork together occupy 17.3\,GiB of GPU memory.

\begin{table}[h]
\centering
\small
\begin{tabular}{lcc}
\toprule
Dataset & GH200 & RTX PRO 6000 \\
\midrule
Amazon23 & 113\,ms & 162\,ms \\
AbsGen   & 62\,ms  & 77\,ms \\
\bottomrule
\end{tabular}
\caption{Averaged per-update latency of \textsc{WASP}$_{\mathrm{cont}}$.}
\label{tab:update-latency}
\end{table}

\section{Dataset Details}
\label{sec:appendix:datasets}
\subsection{Task Details}
\paragraph{LaMP-1: Personalized Citation Identification.}
This is a binary classification task. Given a target paper and two candidate references, the model predicts which candidate the author is more likely to cite. The user history contains the author's previous publications and provides evidence about their research interests and citation preferences.

\paragraph{LaMP-2N: Personalized News Categorization.}
The model assigns a news article to one of 15 categories. The history contains articles previously written by the user together with their category labels, allowing the model to infer the user's topical distribution and labeling patterns.

\paragraph{LaMP-2M: Personalized Movie Tagging.}
Given a movie description, the model selects one of 15 tags. Personalization is based on the user's historical movie--tag pairs, which capture individual differences in how users describe otherwise similar content.

\paragraph{LaMP-3: Personalized Product Rating.}
Given a review, the model predicts the integer rating assigned by the user on a 1--5 scale. The history comprises the user's earlier review--rating pairs and exposes their rating calibration in addition to the sentiment expressed in the current review.

\paragraph{LaMP-4: Personalized News Headline Generation.}
The model generates a headline for a news article. Historical article--headline pairs characterize the user's lexical choices, preferred level of specificity, and headline style.

\paragraph{LaMP-5: Personalized Scholarly Title Generation.}
Given a scientific abstract, the model generates the paper title. The user's previous abstract--title pairs provide domain and stylistic evidence for producing a title consistent with that author.

\paragraph{LaMP-7: Personalized Tweet Paraphrasing.}
The model rewrites an input text as a tweet while preserving its meaning. The user's prior tweets provide evidence about tone, phrasing, and other recurring stylistic choices.

\paragraph{LongLaMP-1: Personalized Abstract Generation.}
Given a paper title and a list of key research items, the model generates the paper's abstract. The user history contains the author's earlier papers, each given as its title and abstract, and therefore tests whether a method can preserve an author's long-form academic style while producing domain-appropriate content.

\paragraph{LongLaMP-2: Personalized Topic Writing.}
Given the summary of a Reddit post, the model generates the full post. The user history contains the author's earlier posts, each given as its summary and full text, and supplies evidence about the author's voice, organization, and typical level of detail.

\paragraph{LongLaMP-3: Personalized Review Writing.}
Given a product description, an overall rating, and a short review summary, the model generates the full review text. The user history contains the user's earlier reviews, each given as the product description and the full review text, requiring the model to capture both writing style and opinion patterns.

\paragraph{Amazon23: Amazon Reviews 2023.}
We build our continual personalized review-generation dataset from Amazon Reviews 2023~\cite{amazon2023}. Following LongLaMP's review-writing formulation~\cite{kumar2024longlamp}, the model generates a full review from the product description, conditioned on the user's previous reviews. Unlike LongLaMP, which uses an earlier Amazon Reviews release, Amazon23 provides newer interactions, richer item metadata, and fine-grained timestamps.

\paragraph{AbsGen: Personalized Abstract Generation.}
We use the temporal version of LongLaMP abstract generation~\cite{kumar2024longlamp}, ordering each author's papers by publication time. At every step, the model generates the abstract of the next paper from its title and research items, conditioned only on earlier publications. This preserves LongLaMP's long-form generation format while turning each author profile into a chronological personalization stream.

\begin{figure*}[t]
  \centering
  \includegraphics[width=\textwidth]{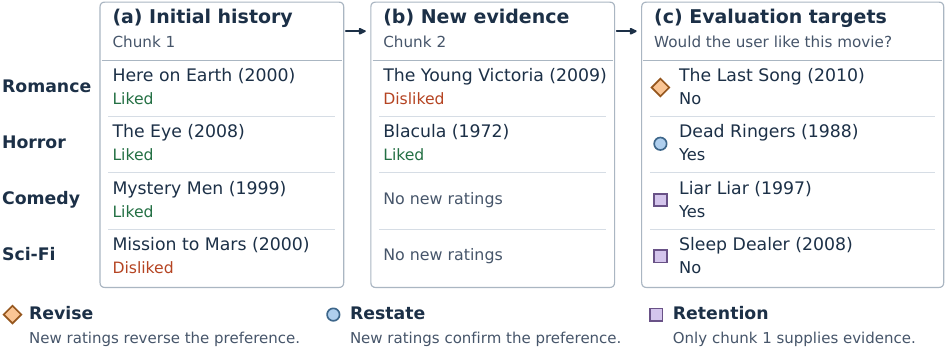}
  \caption{We construct two-chunk episodes from MovieLens 25M with controlled changes to explicit genre preferences. Chunk 1 establishes like/dislike preferences for four genres. Chunk 2 reverses (\emph{revise}) or confirms (\emph{restate}) preferences for selected genres. Omitted genres provide \emph{retention} targets. The example shows one rating per observed genre in each chunk and one held-out movie per genre with its expected answer after the update.}
  \label{fig:appendix:movielens-episode}
\end{figure*}

\subsection{Temporal Data Construction}
For each user, we sort interactions chronologically and split the history into non-overlapping chunks of at most 8 items or 4{,}096 tokens. Each chunk produces one update: after reading $C_1,\ldots,C_t$, the adapter is evaluated by generating the items in the next chunk, $C_{t+1}$. Thus, an item is first a prediction target and later becomes part of the history. We discard users with fewer than two chunks.

We use disjoint user splits, reserving 200 users per task for validation and another 200 for testing. During training, each user's chronologically ordered chunks are divided into non-overlapping episodes of at most 16 consecutive chunks. Within each episode, the model updates the adapter one chunk at a time, and the loss is backpropagated through the full sequence. At each update, the model learns to generate four sampled items from the next chunk. During inference, $C_1$ initializes the adapter and $C_2$ provides a warm-up update; scoring starts at $C_3$ and covers all items in up to 16 chunks.

\subsection{Temporal Analysis of User Histories}
\label{sec:appendix:temporal-analysis}

We examine whether user histories in Amazon23 and AbsGen show evidence of preference shift over time. We first group pairs of reviews or abstracts from the same user by time gap and compute their mean cosine similarity. On both datasets, similarity decreases as the gap grows (Table~\ref{tab:temporal-similarity}). We then test whether the choice of history chunk affects prediction by varying the chunk provided to the base model as in-context evidence for the held-out responses. The user's most recent chunk yields the lowest cross-entropy on both datasets, outperforming even a randomly selected earlier chunk from the same user (Table~\ref{tab:history-recency}). These findings motivate maintaining a user state that tracks relevant preferences as new interactions arrive.

\begin{table}[t]
    \centering
    \small
    \begin{tabular}{@{}lclc@{}}
        \toprule
        \multicolumn{2}{c}{Amazon23} & \multicolumn{2}{c}{AbsGen} \\
        \cmidrule(lr){1-2}\cmidrule(lr){3-4}
        Time gap & Cos.\ sim. & Time gap & Cos.\ sim. \\
        \midrule
        $<$7 days    & .4875 & Same year   & .4233 \\
        7--30 days   & .4691 & 1 year      & .4195 \\
        30--90 days  & .4596 & 2--3 years  & .4087 \\
        90--365 days & .4526 & 4--6 years  & .3906 \\
        1--2 years   & .4469 & 7--10 years & .3749 \\
        $>$2 years   & .4377 & $>$10 years & .3574 \\
        \bottomrule
    \end{tabular}
    \caption{Mean cosine similarity between texts from the same user, grouped by time gap.}
    \label{tab:temporal-similarity}
\end{table}

\begin{table}[t]
    \centering
    \small
    \begin{tabular}{@{}lcc@{}}
        \toprule
        Context Source & Amazon23 & AbsGen \\
        \midrule
        Same user, most recent chunk    & \textbf{2.4512} & \textbf{1.9931} \\
        Same user, random earlier chunk & 2.4742 & 2.0406 \\
        Same user, oldest chunk         & 2.4967 & 2.0536 \\
        Another user's chunk            & 2.5390 & 2.0584 \\
        No context                      & 2.5227 & 2.0464 \\
        \bottomrule
    \end{tabular}
    \caption{Base-model cross-entropy ($\downarrow$) on held-out responses under different history contexts.}
    \label{tab:history-recency}
\end{table}
\subsection{Explicit Preference Dataset Construction}
\label{sec:appendix:dataset:movielens}

We use MovieLens 25M~\cite{harper2015movielens} to construct histories for synthetic users who like or dislike each of four genres (Horror, Comedy, Romance, and Sci-Fi). We select movies associated with a single target genre. Each history entry records the movie title, its genre
labels, and the user's like or dislike judgment.

Each episode consists of two chunks for one user (Figure~\ref{fig:appendix:movielens-episode}). The first chunk establishes the user's initial preferences with two movies per genre. The second covers zero to four genres. For each genre it covers, ratings on two new movies either reverse the initial preference (\emph{revise}) or confirm it (\emph{restate}). Genres omitted from the update provide \emph{retention} targets, for which the initial preference still applies. We fill each chunk to ten movies with randomly labeled titles outside the target genres. After each chunk, we probe each genre with two previously unseen movies. Given a movie's title and genre labels, the model answers \emph{Yes} or \emph{No} when asked whether the user would like it.

Of the $2^4 = 16$ possible initial preference configurations, we use 14 to construct 96000 training episodes and reserve two complementary configurations for 3000 evaluation episodes. Holding out the starting configurations prevents complete preference trajectories from appearing in both splits.

\subsection{Dataset Statistics}
Tables~\ref{tab:p2p_stats} and~\ref{tab:stream_stats} report statistics for the static and continual personalization datasets, respectively. Input and output lengths are measured in characters, whereas Tokens/Chunk is measured in tokens.

\input{tables/tab_p2p_stats}
\input{tables/tab_stream_stats}

\section{Baseline Details}
\label{sec:appendix:baselines}
All baselines use the same task-adapted Qwen2.5-7B-Instruct backbone. For the static experiments, each method receives the same non-chronological user history. In the continual experiments, methods may access only $\mathcal C_{u,1:t}$ when predicting examples in $\mathcal C_{u,t+1}$.

\paragraph{Retrieval-Augmented Generation (\textsc{RAG}).}
Following LaMP and P2P~\cite{salemi2024lamp,p2p}, we use BM25 to retrieve the $k=2$ history items most relevant to the current query and prepend them to the task input. If $R(q_u,\mathcal H_u,k)$ denotes retrieval and $[\cdot\Vert\cdot]$ concatenation, the resulting input is $[R(q_u,\mathcal H_u,k)\Vert q_u]$. No user-specific parameters are trained.

\paragraph{Profile-Augmented Generation (\textsc{PAG}).}
PAG~\cite{PAG} first summarizes a sample of the user's history into a textual profile $s_u$. At inference time, this profile and the same top-$k$ retrieved interactions used by \textsc{RAG} are prepended to the query, giving $[s_u\Vert R(q_u,\mathcal H_u,k)\Vert q_u]$.

\paragraph{Full-History.}
This baseline flattens the complete available history and prepends it to the current query. When the sequence exceeds the backbone's context window, we retain the most recent interactions that fit. In continual evaluation, the prompt contains the accumulated prefix $\mathcal C_{u,1:t}$; for the efficiency analysis, we reuse its key--value cache across updates.

\paragraph{One PEFT Per User (\textsc{OPPU}).}
OPPU~\cite{tan2024oppu} optimizes a separate LoRA adapter on each target user's history while keeping the backbone frozen. Unlike hypernetwork methods, it requires user-specific backpropagation and storage. In the continual setting, we retrain a fresh adapter from the task-adapted model on the complete accumulated prefix at every update, so its update cost grows with history length.

\paragraph{Multi-Task LoRA (\textsc{MT-LoRA}).}
We train a shared LoRA on query--response pairs pooled across training users without user histories, and merge it into the base model.

\paragraph{P2P.}
P2P~\cite{p2p} compresses the available history into a textual profile and maps it to a personalized LoRA. For the efficiency analysis, we adapt it to the continual setting as \textsc{P2P}$_{\mathrm{reappend}}$, which appends each new chunk to the accumulated history, rebuilds the profile, and regenerates the adapter, causing update cost to grow with history length.

\paragraph{SHINE.}
SHINE~\cite{shine} compresses the history into fixed memory tokens and directly generates a LoRA. For the efficiency analysis, we further include \textbf{SHINE-R}, which generates and composes a LoRA for each incoming chunk, causing per-user storage and effective adapter rank to grow with the number of updates.

%% file: tables/tab_personalization.tex
\begin{table*}[t]
\centering
\small
\setlength{\tabcolsep}{5.5pt}
\renewcommand{\arraystretch}{1.05}
\begin{tabular}{lcccccc}
\toprule
\textbf{Metric} & \textbf{Full hist.} & \textbf{RAG}
  & \textbf{PAG} & \textbf{SHINE} & \textbf{P2P} & \textbf{\textsc{WASP} (Ours)} \\
\midrule
Acc $\uparrow$          & .732 & .631 & .635 & \best{.893} & .798 & \sbest{.846} \\
F1 $\uparrow$           & .648 & .547 & .533 & .697 & \sbest{.735} & \best{.786} \\
R-1 $\uparrow$          & .339 & .294 & .306 & \sbest{.376} & .194 & \best{.407} \\
R-L $\uparrow$          & .300 & .283 & .320 & \sbest{.361} & .187 & \best{.383} \\
MAE $\uparrow$          & .915 & .850 & .910 & \sbest{.928} & .905 & \best{.965} \\
Cos. Sim. $\downarrow$  & -- & -- & -- & \sbest{.771} & .998 & \best{.572} \\
All 10 tasks $\uparrow$ & .491 & .444 & .474 & \sbest{.577} & .442 & \best{.580} \\
\bottomrule
\end{tabular}
\caption{On the random split, \textsc{WASP} achieves the highest personalization rate across the ten tasks and the lowest cosine similarity between user adapters. Performance rows report $\text{win\_both}=\Pr(\text{self}\succeq\text{shuffle}\wedge\text{self}\succeq\text{base})$.}
\label{tab:winboth}
\end{table*}

%% file: tables/tab_p2p_stats.tex
\begin{table*}[t]
\centering
\small
\setlength{\tabcolsep}{4pt}
\renewcommand{\arraystretch}{1.05}
\resizebox{\textwidth}{!}{%
\begin{tabular}{lllrrrr}
\toprule
\textbf{Dataset} & \textbf{Task} & \textbf{Split} & \textbf{Users} & \textbf{Queries} & \textbf{Avg. Input Len.} & \textbf{Avg. Output Len.} \\
\midrule
\multirow{21}{*}{LaMP}
 & \multirow{3}{*}{Citation ID}
   & Train       & 5{,}947 & 7{,}289 & 355.25 & 3.00 \\
 & & OOD test    & 200 & 255 & 377.62 & 3.00 \\
 & & Random test & 200 & 254 & 358.93 & 3.00 \\
\cmidrule{2-7}
 & \multirow{3}{*}{Movie Tagging}
   & Train       & 733 & 12{,}515 & 602.82 & 9.46 \\
 & & OOD test    & 93 & 409 & 602.64 & 7.62 \\
 & & Random test & 92 & 518 & 602.92 & 9.36 \\
\cmidrule{2-7}
 & \multirow{3}{*}{News Cat.}
   & Train       & 253 & 13{,}711 & 463.26 & 9.63 \\
 & & OOD test    & 32 & 1{,}106 & 445.29 & 8.90 \\
 & & Random test & 32 & 823 & 449.61 & 9.53 \\
\cmidrule{2-7}
 & \multirow{3}{*}{News Headline Gen.}
   & Train       & 1{,}396 & 12{,}147 & 178.02 & 62.50 \\
 & & OOD test    & 100 & 487 & 219.65 & 49.77 \\
 & & Random test & 100 & 1{,}125 & 174.10 & 63.62 \\
\cmidrule{2-7}
 & \multirow{3}{*}{Product Rating}
   & Train       & 19{,}244 & 21{,}658 & 694.97 & 1.00 \\
 & & OOD test    & 200 & 217 & 493.73 & 1.00 \\
 & & Random test & 200 & 221 & 739.47 & 1.00 \\
\cmidrule{2-7}
 & \multirow{3}{*}{Scholarly Title Gen.}
   & Train       & 14{,}274 & 15{,}738 & 1{,}094.14 & 76.60 \\
 & & OOD test    & 200 & 218 & 1{,}156.29 & 74.27 \\
 & & Random test & 200 & 217 & 1{,}065.66 & 75.35 \\
\cmidrule{2-7}
 & \multirow{3}{*}{Tweet Paraphrase}
   & Train       & 12{,}912 & 14{,}346 & 180.44 & 93.16 \\
 & & OOD test    & 200 & 225 & 184.37 & 93.75 \\
 & & Random test & 200 & 223 & 178.26 & 92.32 \\
\midrule
\multirow{9}{*}{LongLaMP}
 & \multirow{3}{*}{Abstract Generation}
   & Train       & 22{,}103 & 30{,}950 & 261.17 & 1{,}116.49 \\
 & & OOD test    & 200 & 283 & 256.51 & 1{,}098.07 \\
 & & Random test & 200 & 277 & 259.04 & 1{,}106.19 \\
\cmidrule{2-7}
 & \multirow{3}{*}{Product Review}
   & Train       & 15{,}490 & 18{,}928 & 725.36 & 1{,}652.50 \\
 & & OOD test    & 200 & 256 & 745.18 & 1{,}813.55 \\
 & & Random test & 200 & 251 & 740.06 & 1{,}412.50 \\
\cmidrule{2-7}
 & \multirow{3}{*}{Topic Writing}
   & Train       & 15{,}330 & 19{,}897 & 196.90 & 1{,}438.58 \\
 & & OOD test    & 200 & 279 & 184.61 & 1{,}410.62 \\
 & & Random test & 200 & 273 & 208.48 & 1{,}381.02 \\
\bottomrule
\end{tabular}%
}
\caption{Static personalization dataset statistics. Input and output lengths are mean character counts per query.}
\label{tab:p2p_stats}
\end{table*}

%% file: tables/tab_stream_stats.tex
\begin{table*}[t]
\centering
\small
\setlength{\tabcolsep}{4pt}
\renewcommand{\arraystretch}{1.05}
\resizebox{\textwidth}{!}{%
\begin{tabular}{llrrrrrrrr}
\toprule
\textbf{Dataset} & \textbf{Split} & \textbf{Users} & \textbf{Episodes} & \textbf{Updates} & \textbf{Targets} & \textbf{Items/User} & \textbf{Tokens/Chunk} & \textbf{Avg. Input Len.} & \textbf{Avg. Output Len.} \\
\midrule
\multirow{3}{*}{Amazon23}
 & Train & 54{,}848 & 72{,}061 & 539{,}523 & 3{,}840{,}540 & 77.84 & 3{,}101 & 584.84 & 1{,}727.51 \\
 & Valid & 200 & 200 & 2{,}125 & 15{,}549 & 85.57 & 2{,}997 & 558.20 & 1{,}534.89 \\
 & Test  & 200 & 200 & 1{,}750 & 12{,}481 & 70.32 & 3{,}022 & 560.78 & 1{,}713.44 \\
\midrule
\multirow{3}{*}{AbsGen}
 & Train & 22{,}421 & 29{,}641 & 318{,}963 & 2{,}470{,}649 & 118.19 & 1{,}185 & 247.88 & 1{,}000.45 \\
 & Valid & 200 & 200 & 2{,}654 & 20{,}509 & 110.55 & 1{,}172 & 245.16 & 984.01 \\
 & Test  & 200 & 200 & 2{,}733 & 21{,}104 & 113.52 & 1{,}182 & 246.74 & 1{,}001.80 \\
\bottomrule
\end{tabular}%
}
\caption{Continual personalization dataset statistics. Updates count adapter updates, and targets count next-chunk items to predict. Input and output lengths are mean character counts per target.}
\label{tab:stream_stats}
\end{table*}